\pdfoutput=1
\documentclass[11pt]{article}

\usepackage[final]{acl}

\usepackage{tcolorbox}
\usepackage{times}      
\usepackage{latexsym}   

\usepackage[T1]{fontenc}

\usepackage[utf8]{inputenc}

\usepackage{microtype}

\usepackage{inconsolata}

\usepackage{graphicx}

\usepackage{multirow}
\usepackage{makecell} 
\usepackage{xspace}
\usepackage{tcolorbox}

\usepackage{amsmath} 
\usepackage{tcolorbox}
\usepackage{enumitem}
\usepackage{booktabs}
\usepackage{bm}

\def\T#1{{``\textcolor{YellowGreen}{\textit{#1}}''}} 

\makeatletter
\def\UrlAlphabet{%
    \do\a\do\b\do\c\do\d\do\e\do\f\do\g\do\h\do\i\do\j%
    \do\k\do\l\do\m\do\n\do\o\do\p\do\q\do\r\do\s\do\t%
    \do\u\do\v\do\w\do\x\do\y\do\z\do\A\do\B\do\C\do\D%
    \do\E\do\F\do\G\do\H\do\I\do\J\do\K\do\L\do\M\do\N%
    \do\O\do\P\do\Q\do\R\do\S\do\T\do\U\do\V\do\W\do\X%
    \do\Y\do\Z}
\def\UrlDigits{\do\1\do\2\do\3\do\4\do\5\do\6\do\7\do\8\do\9\do\0}
\g@addto@macro{\UrlBreaks}{\UrlOrds}
\g@addto@macro{\UrlBreaks}{\UrlAlphabet}
\g@addto@macro{\UrlBreaks}{\UrlDigits}
\makeatother

\definecolor{deepblue}{HTML}{0000FF} 

\renewcommand{\paragraph}[1]{\smallskip\noindent\textbf{#1.}}
\renewcommand{\subparagraph}[1]{\smallskip\noindent\textbf{\underline{#1.}}}

\title{Don't Reinvent the Wheel: Efficient Instruction-Following Text Embedding based on Guided Space Transformation}

\author{
  \textbf{Yingchaojie Feng\textsuperscript{1}\thanks{Yingchaojie Feng and Yiqun Sun contributed equally to this work.}},
  \textbf{Yiqun Sun\textsuperscript{2}\footnotemark[1]},
  \textbf{Yandong Sun\textsuperscript{2}},
  \textbf{Minfeng Zhu\textsuperscript{3}\thanks{Minfeng Zhu and Qiang Huang are corresponding authors.}},
  \textbf{Qiang Huang\textsuperscript{4}\footnotemark[2]}, \\
  \textbf{Anthony K. H. Tung\textsuperscript{2}},
  \textbf{Wei Chen\textsuperscript{1}} \\
  \textsuperscript{1}State Key Lab of CAD\&CG, Zhejiang University \\
  \textsuperscript{2}School of Computing, National University of Singapore ~~~
  \textsuperscript{3}Zhejiang University \\
  \textsuperscript{4}School of Intelligence Science and Engineering, Harbin Institute of Technology (Shenzhen) \\
  \texttt{\{fycj, minfeng\_zhu, chenvis\}@zju.edu.cn,} \\
  \texttt{\{sunyq, yandong, atung\}@comp.nus.edu.sg}, \texttt{huangqiang@hit.edu.cn}
}

\begin{document}

\maketitle
\begin{abstract}
In this work, we investigate an important task named instruction-following text embedding, which generates dynamic text embeddings that adapt to user instructions, highlighting specific attributes of text.
Despite recent advancements, existing approaches suffer from significant computational overhead, as they require re-encoding the entire corpus for each new instruction. 
To address this challenge, we propose \textbf{GSTransform}, a novel instruction-following text embedding framework based on \textbf{G}uided \textbf{S}pace \textbf{Transform}ation.
Our key observation is that instruction-relevant information is inherently encoded in generic embeddings but remains underutilized.
Instead of repeatedly encoding the corpus for each instruction, GSTransform is a lightweight transformation mechanism that adapts pre-computed embeddings in real time to align with user instructions, guided by a small amount of text data with instruction-focused label annotation.
We conduct extensive experiments on three instruction-awareness downstream tasks across nine real-world datasets, demonstrating that GSTransform improves instruction-following text embedding quality over state-of-the-art methods while achieving dramatic speedups of 6$\sim$300$\times$ in real-time processing on large-scale datasets.
The source code is available at \url{https://github.com/YingchaojieFeng/GSTransform}.
\end{abstract}

\section{Introduction}
\label{sec:intro}

\begin{figure*}[t]
  \centering
  \includegraphics[width=0.99\textwidth]{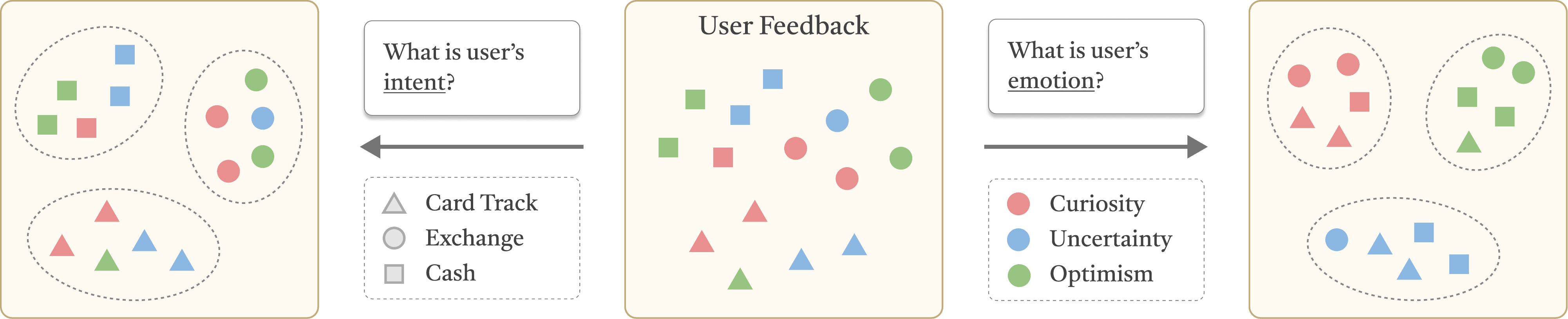}
  \vspace{-0.5em}
  \caption{An illustration of instruction-following text embeddings. Depending on different instructions, the embeddings reorganize their similarity relationships, forming distinct clusters that emphasize different aspects.}
  \label{fig:illustration}
  \vspace{-0.75em}
\end{figure*}

Text embedding \citep{NIPS2013_9aa42b31, pennington-etal-2014-glove, zhuo-etal-2023-whitenedcse, li-li-2024-aoe} is a fundamental NLP problem, serving as the backbone of numerous applications such as clustering \citep{aggarwal2012survey}, Semantic Textual Similarity (STS) \citep{agirre-etal-2012-semeval, agirre-etal-2013-sem}, and information retrieval \citep{karpukhin-etal-2020-dense, thakur2beir}.
The core objective of text embedding models is to convert textual data into fixed-length vector representations that encode semantic relationships, ensuring that similar texts remain close in the embedding space while dissimilar ones are well separated \citep{muennighoff-etal-2023-mteb}.

Despite their widespread use, generic embedding models are inherently static and inflexible, often failing to adapt to task-specific requirements.
As illustrated in Figure~\ref{fig:illustration}, real-world text data frequently encapsulates multiple semantic aspects (e.g., ``intent'' and ``emotion''), yet traditional embeddings primarily focus on general semantics rather than user-specified perspectives.
This limitation restricts their applicability in scenarios where users seek embeddings tailored to specific needs.

To address this challenge, recent research has introduced instruction-following text embedding, which generates embeddings in different semantic spaces conditioned on user-provided instructions \citep{su-etal-2023-one, peng-etal-2024-answer}. 
These models allow users to emphasize specific aspects of text, dynamically adjusting embeddings based on task-oriented instructions.
Notably, InstructOR \cite{su-etal-2023-one} concatenates instruction with text as input and fine-tunes a Transformer-based model across a diverse set of instructions to capture instruction-aware semantics. 
Alternatively, InBedder \cite{peng-etal-2024-answer} treats instructions as questions and encodes the corresponding answers as the final embeddings, leveraging both strengths of generative models and embedding models.

Although these methods improve instruction-awareness, they suffer from significant efficiency limitations when applied to large-scale datasets.
The primary issue stems from their dependence on full corpus re-encoding for each instruction. 
Given $m$ instructions and $n$ texts, both InstructOR and InBedder require $O(m \times n)$ forward passes to generate embeddings, making it computationally expensive and impractical for real-world applications where pre-encoded embeddings are stored in vector databases.
This inefficiency hinders experimentation with different instructions, as even minor changes require reprocessing the entire dataset.

On the other hand, we observe that generic text embeddings already contain latent instruction-relevant information but do not prioritize it explicitly.
Instead of re-encoding the full corpus for every new instruction, we propose a transformation-based approach that dynamically adjusts existing embeddings to align with user-specified instructions.
As illustrated in Figure~\ref{fig:illustration}, different users may prioritize distinct semantic attributes.
For instance, when the focus is on the ``intent'' aspect, the transformation highlights the shape of embeddings to cluster texts based on their underlying purpose. 
Conversely, when the emphasis shifts to the ``emotion'' aspect, the transformation accentuates color, grouping texts by their sentiment.

To this end, we introduce \textbf{GSTransform}, a novel instruction-following text embedding framework based on \textbf{G}uided \textbf{S}pace \textbf{Transform}ation, eliminating the need for exhaustive re-encoding.
GSTransform consists of two components: 
\begin{itemize}[nolistsep]
  \item \textbf{Instruction-based Label Construction:}
  We build a label taxonomy from user instructions to categorize text representations according to instruction-specific semantics.
  
  \item \textbf{Label-guided Embedding Transformation:}
  We adapt the original embedding space using these instruction-driven labels to align with user-specified information.
\end{itemize}
Rather than regenerating embeddings from scratch, GSTransform treats instructions as transformation operators that strategically reorient attention within the existing semantic space, ensuring adaptability without excessive computation.
The instruction-driven label data guides the training of transformation models, providing a clearer optimization objective than direct training on instructions.

\paragraph{Contributions}
This work makes the following essential contributions:
\begin{enumerate}[nolistsep,label*=(\arabic*)]
  \item \textbf{Efficient Instruction-Following Embedding:} 
  We propose GSTransform, a novel framework that enables instruction-following text embeddings via guided space transformation, eliminating the need for re-encoding the entire corpus for each instruction.
  
  \item \textbf{Cost-Effective Instruction Adaptation:} 
  GSTransform introduces a Label-guided Embedding Transformation mechanism that adapts pre-computed embeddings to user instructions using a lightweight model trained on a small, annotated subset, requiring only a fixed number of LLM calls, making it highly cost-effective for large-scale applications.
  
  \item \textbf{Comprehensive Empirical Validation:} 
  We conduct extensive experiments on three instruction-awareness tasks across nine real-world datasets, showing that GSTransform improves embedding quality while significantly reducing computational overhead and latency on large-scale datasets compared to state-of-the-art baselines.
\end{enumerate}
\section{Related Work}
\label{sec:related_work}

\subsection{Generic Text Embedding}
Text embedding has been a long-studied problem. Since word embeddings, people adopt self-supervised training in generating word embeddings, and pool the word embeddings to form text embeddings \cite{NIPS2013_9aa42b31, pennington-etal-2014-glove}. Recent advancements in context-aware semantic text embedding models leverage Transformer-based architectures \cite{transformer, devlin-etal-2019-bert} as their backbone, often employing customized objectives like contrastive loss to train the models \cite{cer-etal-2018-universal, reimers-gurevych-2019-sentence, gao-etal-2021-simcse, zhuo-etal-2023-whitenedcse}. Moreover, state-of-the-art (SOTA) text embedding models have been further enhanced with techniques such as using large language models (LLMs) \cite{wang2023improving, muennighoff2024generative, lei-etal-2024-meta} and more sophisticated loss functions designed to address issues like cosine saturation \cite{li-li-2024-aoe}. 

Despite their effectiveness, these methods lack generalizability and fail to meet diverse user needs when downstream tasks require focusing on specific aspects beyond general semantics.

\subsection{Instruction-Following Text Embedding}
Instruction-following text embedding~\cite{su-etal-2023-one, peng-etal-2024-answer} allows users to guide embedding generation through customized instructions. 
The model produces embeddings that align with users' specific interests by considering both the input text and instructions.

InstructOR~\cite{su-etal-2023-one} pioneered instruction-based embeddings by concatenating instructions with input texts and training the model using contrastive objectives across a diverse set of instructions. 
It adapts embeddings for varied semantic interpretations but does not explicitly model instruction-specific semantic aspects. 
InBedder~\cite{peng-etal-2024-answer} extends this idea by treating instructions as questions and generating intermediate answers to produce more fine-grained, instruction-aware embeddings. 
They also propose Instruction Awareness Tests, which we adopt to evaluate Triplet Alignment, STS, and Clustering tasks.
Yet, both methods require re-encoding the entire corpus for each new instruction, resulting in notable computational overhead and latency, especially for large-scale datasets.

Beyond text embeddings, related efforts have explored instruction-aware and prompt-based information retrieval~\cite{weller2024promptriever, min2024unihgkr, oh2024instructir, sun2024mair, weller2024followir}, offering alternative formulations that leverage user intent to enhance retrieval quality.

\section{The GSTransform Framework}
\label{sec:framework}

\begin{figure*}[t]
\centering
\includegraphics[width=0.99\linewidth]{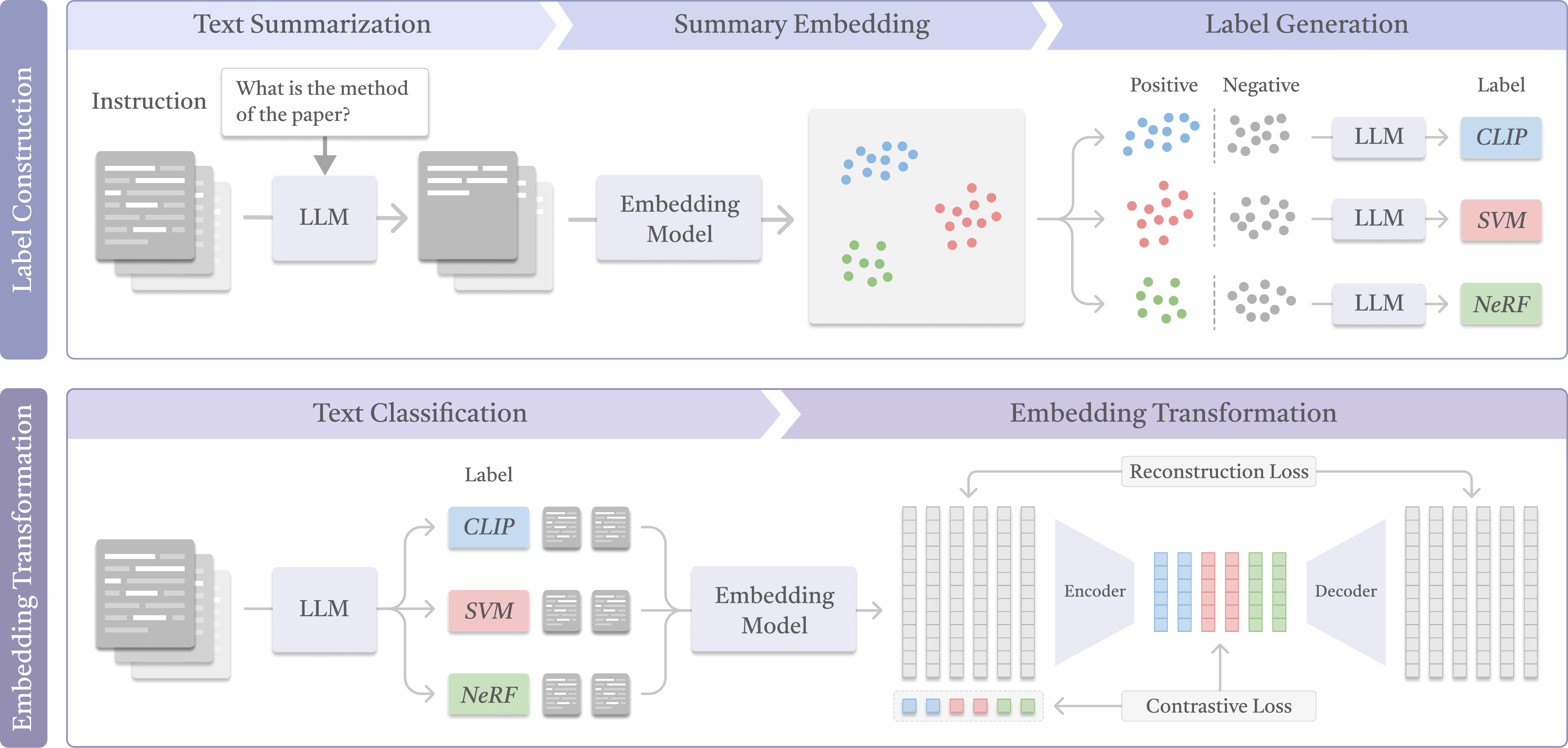}
\caption{GSTransform consists of two core components: (1) Instruction-based Label Construction, which summarizes the texts to extract instruction-relevant information, performs embedding clustering on summary text, and generates labels to represent their characteristics; (2) Label-Guided Embedding Transformation, which classifies the text based on the constructed labels and transforms the generic embedding space guided by the labeled data.}
\label{fig:framework}
\vspace{-0.5em}
\end{figure*}

We introduce GSTransform, a novel framework designed for efficient instruction-following embedding generation, enabling dynamic adaptation of text embeddings to user-specified instructions.
In contrast to prior methods that require re-encoding the entire corpus for every instruction, GSTransform operates directly on pre-computed embeddings, allowing real-time transformation with minimal computational overhead. 

As illustrated in Figure \ref{fig:framework}, GSTransform comprises two core components: 
\begin{enumerate}[nolistsep,label*=(\arabic*)]
  \item \textbf{Instruction-based Label Construction:} 
  This component constructs an instruction-aware label taxonomy to guide downstream transformation. 
  A small corpus of texts is randomly sampled, and instruction-guided summaries are generated using an LLM. 
  These summaries help to capture the key semantic aspects emphasized by the instruction, enabling the taxonomy to reflect fine-grained, instruction-relevant distinctions.
  \item \textbf{Label-guided Embedding Transformation:} 
  Once the label taxonomy is constructed, we use an LLM to annotate the sampled texts according to the taxonomy.
  The resulting labeled data is then used to train a lightweight Embedding Transformation Model, which maps generic embeddings to an instruction-aligned semantic space. 
  The model jointly optimizes a contrastive loss, which is used to highlight instruction-specific similarities, and a reconstruction loss, which aims to preserve general semantic structure.
\end{enumerate}

In all LLM-involved stages (summarization, labeling, classification), we use GPT-4o-mini for its balance of performance and efficiency, and the prompt templates are provided in Appendix \ref{appendix:prompts}.
By decoupling instruction adaptation from embedding generation, GSTransform supports scalable, real-time transformations without the need for repeated re-encoding.
We now begin by detailing the first component: Instruction-based Label Construction.

\subsection{Instruction-based Label Construction}
\label{sec:framework:label-construction}

To ensure that the label taxonomy aligns with user-specified instructions while remaining sensitive to dataset-specific nuances, we adopt a bottom-up construction strategy. 
Rather than processing the entire corpus, which is often computationally expensive for large corpora, we randomly sample a representative subset (e.g., 3,000 texts in our experiments) for label construction.
While instruction-aware labeled data can be pre-defined, this component addresses the common scenario where such labeled data is unavailable or costly to obtain.

\paragraph{Instruction-Following Clustering}
Traditional label construction methods, such as BERTopic \cite{grootendorst2022bertopic}, typically involve clustering embeddings obtained from generic models,  followed by heuristic label generation.
However, these methods often fail to capture instruction-specific distinctions, as generic embeddings do not prioritize user-specified attributes.

To overcome this limitation, we introduce Instruction-following Clustering, a method designed to extract and amplify instruction-relevant semantics from the input text.
The process comprises three key steps:
\begin{itemize}[nolistsep]
  \item \textbf{Step 1: Instruction-Guided Summarization.} 
  We first use an LLM to generate concise summaries of each sampled text, tailored to the user's instructions.
  
  \item \textbf{Step 2: Embedding with Generic Models.} 
  The instruction-guided summaries are then encoded using a generic embedding model, such as UAE (Universal AnglE Embedding) \cite{li-li-2024-aoe}.
  
  \item \textbf{Step 3: Clustering.} 
  The resulting summary embeddings are grouped using the $k$-means$++$ algorithm \cite{kmeanspp}, producing clusters that align with instruction-specified aspects.
\end{itemize}

The number of clusters ($k$) determines the granularity of the label taxonomy. 
We set $k=50$ as a default, which we found to offer a good balance between semantic specificity and generalizability without extensive hyperparameter tuning.
The effect of $k$ is studied in Section~\ref{sec:expt:params}.

\paragraph{Label Generation}
To semantically characterize each cluster, we generate representative labels using LLMs.
However, due to the non-deterministic nature of LLM outputs, the generated labels may vary in quality, ranging from \emph{overly generic} (failing to distinguish between clusters) to \emph{overly specific} (lacking generalizability beyond the sampled texts).

To address this, we adopt a contrastive prompting strategy inspired by principles in contrastive learning. 
Specifically, each LLM prompt is composed of both \emph{positive examples} (texts sampled from within the target cluster) and \emph{negative examples} (texts drawn from other clusters), encouraging the LLM to focus on the discriminative features that define the cluster while being aware of what sets it apart from others.

This strategy ensures that the generated labels are: semantically aligned with the instruction-relevant focus, mutually exclusive across clusters, and generalizable to unseen examples beyond the sampled subset.
As a result, the generated labels serve as reliable supervision signals for training the embedding transformation model in the next stage.

\subsection{Label-Guided Embedding Transformation}
\label{sec:framework:embed-transform}

This component transforms the original embedding space into a new space that focuses on instruction-specific information. 
We first annotate the sample text using the generated label taxonomy and then train a transformation model to project generic embeddings into an instruction-adapted space.

\paragraph{Text Classification}
In this step, we use the generated label taxonomy to annotate sample data for training the transformation model.
Given the powerful capability of LLMs in semantic understanding and instruction following, we employ LLMs to classify the sampled text based on user instructions and the generated label taxonomy.
We classify the original text instead of the summary to leverage the rich context of the original text.

\paragraph{Embedding Transformation}
The transformation model aims to restructure the embedding space to emphasize instruction-relevant features while preserving general semantic integrity.
Once trained, it can dynamically transform any embedding from the generic space into an instruction-specific space, enabling efficient large-scale processing without requiring access to the original texts.

Specifically, the transformation model adopts a lightweight encoder-decoder architecture, where both the encoder and decoder are implemented as \emph{single-layer linear models}. 
This simple yet effective design avoids overfitting and ensures that the transformation preserves the overall structure of the original embedding space.

The encoder transforms the each input embedding vector ($\bm{x}_i$) into an instruction-aware representation ($\bm{e}_i$), while the decoder reconstructs the original embedding ($\bm{\hat{x}}_{i}$) from the transformed vector ($\bm{e}_i$).
The model is trained with a dual-objective loss function that balances semantic adaptation and information preservation:
\begin{displaymath}
  \mathcal{L} = \beta_1 \cdot \mathcal{L}_{contr} + \beta_2 \cdot \mathcal{L}_{recon},
\end{displaymath}
where $\beta_1$ and $\beta_2$ are weight coefficients (both set to 1.0 by default).

The \textbf{contrastive loss} $\mathcal{L}_{contr}$ encourages embeddings of texts sharing the same label (as defined by the instruction-based taxonomy) to be closer, while separating those with different labels:
\begin{align*}
  \mathcal{L}_{contr} = \frac{1}{|N|^2} \sum_{i, j \in N} \Big( 
  & \mathbf{1}_{y_i = y_j} \cdot D(\bm{e}_i, \bm{e}_j)^2 + \\
  & \mathbf{1}_{y_i \neq y_j} \cdot D'(\bm{e}_i, \bm{e}_j)^2 \Big),
\end{align*}
where $D$ represents the Euclidean distance, and $D'(\bm{e}_i, \bm{e}_j) = \max(0, m - D(\bm{e}_i, \bm{e}_j))$ incorporates a margin 
$m$ to enforce separation between classes. 
$\mathbf{1}_{y_i = y_j}$ is an indicator that equals 1 if $\bm{e}_i$ and $\bm{e}_j$ belong to the same class and 0 otherwise. 

The \textbf{reconstruction loss} $\mathcal{L}_{recon}$ minimizes information loss by penalizing the differences between the reconstructed $\bm{\hat{x}}_{i}$ and original embeddings $\bm{x}_{i}$:
\begin{displaymath}
  \mathcal{L}_{recon} = \frac{1}{N} \sum_{i=1}^{N} {\Vert \bm{\hat{x}}_{i} - \bm{x}_{i} \Vert}^2.
\end{displaymath}

To train the model, we use the instruction-annotated subset (e.g., 3,000 samples), splitting it into 80\% training and 20\% validation.
We apply early stopping based on validation loss to ensure efficient convergence and avoid overfitting.
Once trained, the model can be applied to transform any generic embedding into an instruction-aligned representation, enabling scalable and real-time adaptation without requiring access to the original text.
\section{Experiments}
\label{sec:expt}

In this section, we conduct a comprehensive evaluation of GSTransform to assess its effectiveness, efficiency, and robustness across a range of instruction-aware embedding tasks. 
In particular, our experiments are designed to answer the following research questions (RQs):
\begin{itemize}[nolistsep]
  \item \textbf{RQ1 (Effectiveness):} Can GSTransform improve the instruction-following capability of generic embedding models across diverse tasks and datasets? (Section~\ref{sec:expt:main})
  
  \item \textbf{RQ2 (Efficiency):} How does GSTransform compare to state-of-the-art instruction-aware embedding baselines in terms of runtime and cost?  (Section~\ref{sec:expt:efficiency})

  \item \textbf{RQ3 (Component Contribution):} What is the contribution of each individual component in the GSTransform framework? (Section~\ref{sec:expt:ablation})

  \item \textbf{RQ4 (Robustness):} How robust is GSTransform to variations in hyperparameters, such as sample size and clustering granularity? (Section~\ref{sec:expt:params}) 
\end{itemize}
To complement the quantitative results, we also include a Case Study (Section~\ref{sec:expt:case}) to qualitatively illustrate how GSTransform adapts embedding structures to different instruction foci.

\subsection{Baseline Models}
\label{sec:expt:baselines}

In the experiments, we evaluate GSTransform on three prevalent embedding models: \textbf{UAE} (Universal AnglE Embedding) \cite{li-li-2024-aoe}, \textbf{Mxbai} \cite{emb2024mxbai}, and \textbf{BGE} (BAAI General Embedding) \cite{bge_embedding}.
These models, designed for generic embeddings, inherently lack instruction-following capabilities.

To assess the impact of GSTransform, we examine how well it enhances their ability to follow user instructions. 
For comparison, we benchmark GSTransform against three state-of-the-art instruction-aware baselines: \textbf{InstructOR} (\texttt{instructor-large}) \cite{su-etal-2023-one}, \textbf{InBedder-Roberta} (\texttt{roberta-large-InBedder}), and \textbf{InBedder-Llama2} (\texttt{llama-2-7b-InBedder}) \cite{peng-etal-2024-answer}.
For a fair evaluation, we use default open-source configurations for all models.

\subsection{Downstream Tasks and Datasets}
\label{sec:expt:datasets}

To evaluate embedding performance in an instruction-aware setting, we follow the protocol established in InBedder \cite{peng-etal-2024-answer} and assess GSTransform on three representative downstream tasks: Clustering, Semantic Textual Similarity (STS), and Triplet Alignment.
For each task, we employ three diverse datasets, each capturing either single or multi-dimensional semantic attributes.
For instance, the Papers with Codes (PaperCode) dataset \cite{ostendorff2022specialized} contains both a \emph{task} and a \emph{method} dimension, enabling evaluation from multiple perspectives.

We craft user instructions to explicitly highlight the target dimensions.
The instructions for baselines follow their default stylistic conventions: command-style instructions for InstructOR and question-style for InBedder. Both styles preserve the same semantic intent.
GSTransform also uses question-style instructions.
Full task and dataset details are summarized in Appendix~\ref{appendix:tasks_datasets}, and the corresponding instructions are detailed in Appendix~\ref{appendix:instructions}.
Below, we briefly describe the three tasks.

\paragraph{Clustering}
This task evaluates how well the structure of the embedding space aligns with the semantic focus of the instruction.
We assess clustering performance on three datasets: NYTClustering (\textbf{NYTClust})~\cite{peng-etal-2024-answer}, Amazon CounterFactual (\textbf{AmzCF})~\cite{o2021wish}, and MasakhaNews (\textbf{MNews})~\cite{adelani2023masakhanews}.
Clustering quality is measured using the \textbf{V-measure} \cite{rosenberg2007v}, which quantifies the consistency between predicted clusters and ground-truth labels.
For NYTClust, which includes multiple aspects, we report the harmonic mean across aspect-specific results.

\paragraph{Semantic Text Similarity (STS)}
This task examines whether embeddings of sentence pairs correctly reflect their semantic similarity under the instruction-defined perspective.
We use three datasets for this task: \textbf{PaperCode} \cite{ostendorff2022specialized}, Multi-HateCheck (\textbf{MultiHate}) \cite{rottger-etal-2022-multilingual}, and \textbf{Big Patent} \cite{sharma2019bigpatent}.
Following \cite{peng-etal-2024-answer}, we sample 50,000 sentence pairs per instruction from annotated datasets. 
Each pair is then assigned a binary label (1 if they share the same class, 0 otherwise) \cite{peng-etal-2024-answer}.
Finally, we compute \textbf{Spearman correlation} between the binary labels and the cosine similarities of the embedding pairs.
Since the Spearman correlation can be negative, we report the arithmetic mean for datasets with multiple aspects.

\paragraph{Triplet Alignment}
This task evaluates whether the embeddings capture relative similarity across three texts: an anchor, a positive, and a negative. 
A correct alignment requires that the distance between the anchor and positive be smaller than that between the anchor and negative.
We use three datasets for this task: IntentEmotion (\textbf{IntEmo})~\cite{peng-etal-2024-answer}, Toxic\_conversations\_50k (\textbf{Toxic})~\cite{do2019jigsaw}, and \textbf{AG-News} \cite{zhang2015character}.
For each dataset, we randomly sample 50,000 triplets, where the anchor and positive share the same label, and the negative has a different label.
We report \textbf{Triplet Alignment Accuracy} as the proportion of triplets where this relative ordering is correctly captured. 
For multi-aspect datasets like IntEmo, we report the harmonic mean across aspects.

\begin{table*}[ht]
\small
\renewcommand{\arraystretch}{1.1}
\setlength\tabcolsep{4pt}
\centering 
\resizebox{\textwidth}{!}{
\begin{tabular}{l rrr rrr rrr r}
  \toprule
  \multirow{2.5}{*}{\textbf{Model}}
  & \multicolumn{3}{c}{\textbf{Clustering (V-measure $\uparrow$)}} & \multicolumn{3}{c}{\textbf{STS (Spearman Corr. $\uparrow$)}} & \multicolumn{3}{c}{\textbf{Triplet Alignment (Accu. $\uparrow$)}} & \\ \cmidrule(lr){2-4} \cmidrule(lr){5-7} \cmidrule(lr){8-10}
  & \textbf{NYTClust} & \textbf{AmzCF} & \textbf{MNews} & \textbf{PaperCode} & \textbf{MultiHate} & \textbf{Big Patent} & \textbf{IntEmo} & \textbf{Toxic} & \textbf{AG-News} & \multirow{-2.5}{*}{\textbf{Mean $\uparrow$}} \\ 
  \midrule
  \textbf{InstructOR} & 53.88 & 1.49 & 60.33 & 55.53 & 36.46 & 21.99 & 51.96 & 53.75 & 73.41 & 45.42 \\
  \textbf{InBedder-Roberta} & 58.47 & 0.40 & 45.17 & 43.82 & 33.27 & 24.44 & 92.64 & 51.49 & 80.99 & 47.85 \\
  \textbf{InBedder-Llama2} & 72.70 & 1.35 & 61.14 & 46.65 & 45.67 & 37.81 & 90.18 & 55.41 & 86.93 & 55.31 \\ 
  \midrule
  \textbf{UAE} & 52.72 & 1.63 & 62.24 & 64.53 & 37.73 & 21.64 & 36.25 & 55.20 & 72.59 & 44.95 \\
  \textbf{GSTransform (UAE)} & 73.66 & \textbf{34.68} & 63.47 & 83.14 & 53.17 & 38.05 & 96.72 & 63.01 & 87.13 & 65.89 \\
  \textbf{Relative Gain} & \textcolor{deepblue}{+20.94} & \textcolor{deepblue}{+33.05} & \textcolor{deepblue}{+1.23} & \textcolor{deepblue}{+18.61} & \textcolor{deepblue}{+15.44} & \textcolor{deepblue}{+16.41} & \textcolor{deepblue}{+60.46} & \textcolor{deepblue}{+7.81} & \textcolor{deepblue}{+14.54} & \textcolor{deepblue}{+20.94} \\
  \midrule
  \textbf{Mxbai} & 54.95 & 0.13 & 63.09 & 64.85 & 38.10 & 22.84 & 35.37 & 55.67 & 75.41 & 45.60 \\
  \textbf{GSTransform (Mxbai)} & 73.92 & 34.14 & \textbf{64.71} & \textbf{83.43} & 52.45 & \textbf{38.34} & 96.30 & \textbf{63.53} & \textbf{87.27} & \textbf{66.01} \\
  \textbf{Relative Gain} & \textcolor{deepblue}{+18.98} & \textcolor{deepblue}{+34.01} & \textcolor{deepblue}{+1.62} & \textcolor{deepblue}{+18.58} & \textcolor{deepblue}{+14.35} & \textcolor{deepblue}{+15.50} & \textcolor{deepblue}{+60.94} & \textcolor{deepblue}{+7.86} & \textcolor{deepblue}{+11.86} & \textcolor{deepblue}{+20.41} \\
  \midrule
  \textbf{BGE} & 55.90 & 0.13 & 47.09 & 65.04 & 37.84 & 16.16 & 42.83 & 54.88 & 71.46 & 43.48 \\
  \textbf{GSTransform (BGE)} & \textbf{75.60} & 33.04 & 63.35 & 83.13 & \textbf{53.24} & 34.56 & \textbf{96.75} & 61.20 & 87.01 & 65.32 \\
  \textbf{Relative Gain} & \textcolor{deepblue}{+19.69} & \textcolor{deepblue}{+32.91} & \textcolor{deepblue}{+16.26} & \textcolor{deepblue}{+18.09} & \textcolor{deepblue}{+15.40} & \textcolor{deepblue}{+18.40} & \textcolor{deepblue}{+53.93} & \textcolor{deepblue}{+6.32} & \textcolor{deepblue}{+15.55} & \textcolor{deepblue}{+21.84} \\
  \bottomrule
\end{tabular}}
\vspace{-0.5em}
\caption{Experimental results on three downstream tasks with their nine associated datasets.}
\label{tab:results}
\vspace{-0.5em}
\end{table*}
\setlength{\textfloatsep}{1.0em}

\subsection{Main Results}
\label{sec:expt:main}

Table~\ref{tab:results} presents the performance comparison results of GSTransform against baseline models.
The strongest baseline, InBedder-Llama2, achieves an average score of 55.31 across the datasets.
In contrast, GSTransform significantly outperforms all baselines, achieving an average score of 66.01, demonstrating its effectiveness in improving instruction-following embeddings.
A particularly notable gain is observed on the AmzCF dataset, where GSTransform improves performance from 1.49 to 34.68, highlighting its ability to extract instruction-relevant information from generic embeddings. 
These results suggest that our approach effectively enhances the instruction-aware clustering quality by aligning the embedding space with user-specific instructions.

Moreover, GSTransform improves the instruction following capability of generic embedding models such as UAE, Mxbai, and BGE. 
For instance, applying GSTransform to BGE enhances performance on NYTClust from 55.90 to 75.60. 
The consistent gains across different backbone models demonstrate the robustness of our approach in enhancing instruction-aware embeddings.

We also notice that the performance gains of GSTransform vary across tasks. We believe this can be partially attributed to the degree of alignment between the datasets and pretraining data of (large) language models. 
For widely used datasets like AG-News and NYTClust, baseline models have likely been pre-trained on similar or overlapping content, leaving limited headroom for further improvements.
In contrast, datasets such as Big Patent involve domain-specific terminology and complex patent structures that general-purpose LLMs may not represent well. This semantic mismatch can reduce the effectiveness of instruction-based adaptation, resulting in smaller relative gains.

\begin{table*}[ht]
\small
\renewcommand{\arraystretch}{1.1}
\setlength\tabcolsep{4pt}
\centering 
\resizebox{\textwidth}{!}{
\begin{tabular}{lrrrrrr}
  \toprule
  \multirow{2.5}{*}{\textbf{Model}} & \multicolumn{3}{c}{\textbf{AG-News}} & \multicolumn{3}{c}{\textbf{Big Patent}} \\ \cmidrule(lr){2-4} \cmidrule(lr){5-7}
  & \textbf{Pre. Time (s) $\downarrow$} & \textbf{RT-Latency (s) $\downarrow$} & \textbf{Est. Cost (\$) $\downarrow$} & \textbf{Pre. Time (s) $\downarrow$} & \textbf{RT-Latency (s) $\downarrow$} & \textbf{Est. Cost (\$) $\downarrow$} \\
  \midrule
  \textbf{InstructOR} & - & 511 & 0.43 & - & 1,679 & 1.40 \\ 
  \textbf{InBedder-Roberta} & - & 749 & 0.62 & - & 1,781 & 1.48 \\ 
  \textbf{InBedder-Llama2} & - & 16,206 & 13.51 & - & 29,756 & 24.80 \\ 
  \textbf{GSTransform} & 626 & 77 & 0.96 & 1,689 & 87 & 2.37 \\
  \bottomrule
\end{tabular}}
\caption{Computational efficiency results on two large-scale datasets. We report the Pre-computing Time (Pre. Time), Real-Time Latency (RT-Latency, per instruction), and Estimated Cost (for 3 instructions) for each model.}
\label{tab:efficiency}
\vspace{-0.5em}
\end{table*}

\subsection{Computational Efficiency}
\label{sec:expt:efficiency}

To assess computational efficiency, we compare GSTransform with the three baselines on two large-scale datasets (AG-News and Big Patent) in terms of time and cost.
AG-News contains 127.6K texts with character lengths ranging from 100 to 1010, while Big Patent consists of 67.1K texts with lengths varying between 2.76K to 3.11M.
Big Patent comprises 67.1K text with character lengths ranging from 2.76K to 3.11M.
All models are evaluated on a server equipped with an NVIDIA A100 GPU (80GB memory) to ensure fairness.

\paragraph{Time and Cost Efficiency}
For time evaluation, we measure the \emph{embedding generation time} for three different instructions. 
Unlike baselines that require re-encoding the entire dataset per instruction, GSTransform supports one-time pre-computing generic embeddings and applying real-time transformations, significantly reducing latency.
Thus, we separately calculate the time for pre-computing and real-time transformation.
For cost evaluation, we calculate the server rental cost (1 dollar per hour) and OpenAI API call costs.

\paragraph{Result Analysis}
Table \ref{tab:efficiency} presents the results:
Baseline models exhibit scalability issues, as their real-time latency is directly proportional to dataset size. Notably, InBedder-Llama2 suffers from severe inefficiencies, requiring 29,756 seconds due to its question-answering generation pipeline.
In contrast, GSTransform processes only a small annotated subset (3,000 texts), reducing real-time latency to just 87 seconds, making it 6$\sim$300$\times$ faster than existing methods.
We also compare the real-time latency of these models at different scales. The results (see Appendix~\ref{app:efficiency_comparison}) show baseline models exhibit latency that increases proportionally with data size, making them more suitable for small-scale datasets. In contrast, GSTransform maintains near-constant latency, since its transformation is learned from a fixed sample size. This results in clear efficiency gains on large-scale datasets.

In terms of cost, GSTransform achieves state-of-the-art performance at a cost comparable to smaller models, making it a scalable and economical solution for real-world applications. Additionally, its flexibility enables seamless integration with future advanced and cost-efficient LLMs.

\subsection{Ablation Studies}
\label{sec:expt:ablation}

We conduct ablation studies to assess the contributions of key components in GSTransform.
For Instruction-based Label Construction, we design two alternative solutions: (1) Removing the instruction-based text summarization component and (2) Replacing the entire process with directed LLM-based label generation.
For Label-guided Embedding Transformation, we replace the transformation model with Fisher Discriminant Analysis (FDA) \cite{mika1999fisher}.
We also assess different embedding backbones to test the generalizability of our method.
The results for three datasets are shown in Table \ref{tab:ablation}, while full results for all datasets are provided in Appendix \ref{app:full_ablation}.

\begin{table}[t]
\renewcommand{\arraystretch}{1.1}
\setlength\tabcolsep{4pt}
\centering 
\resizebox{\columnwidth}{!}{
\begin{tabular}{lccccc}
  \toprule
  \multirow{2.5}{*}{\textbf{Solutions}} & \multicolumn{3}{c}{\textbf{Dataset}} & \multirow{2.5}{*}{\textbf{Mean $\uparrow$}} \\ \cmidrule{2-4} 
  & \textbf{NYTClust} & \textbf{MultiHate} & \textbf{IntEmo} &  \\ \midrule
  \textbf{GSTransform (Mxbai)} & 73.92 & 52.45 & 96.30 & 74.22 \\
  \textbf{Remove Summ.} & 68.32 & 16.21 & 85.14 & 56.56 \\
  \textbf{Directed Label Gen.} & 42.50 & 20.55 & 86.76 & 49.94 \\
  \textbf{FDA-based Transf.} & 67.53 & 48.49 & 89.19 & 68.40 \\
  \bottomrule
\end{tabular}}
\caption{The performance comparison of different solutions in ablation studies.}
\label{tab:ablation}
\end{table}

\paragraph{Remove Summarization}
Removing instruction-based text summarization leads to a significant performance drop (74.22 $\rightarrow$ 56.56), particularly affecting datasets like MultiHate. 
Further analysis reveals that generic embeddings \emph{fail} to distinguish between hateful and non-hateful texts, causing the non-hateful category to be overwhelmed in clustering, reinforcing the importance of instruction-guided summarization.

\paragraph{Directed Label Generation} 
Eliminating the text summarization, summary clustering, and label generation steps and directly relying on LLMs for label generation results in an even steeper performance decline (74.22 $\rightarrow$ 49.94).
We observe that LLM-generated labels often fail to align with dataset characteristics, leading to suboptimal text classification and embedding transformations. 
This suggests that context-aware taxonomy construction is crucial for effective instruction adaptation.

\paragraph{FDA-based Transformation} 
Replacing the encoder-decoder architecture with FDA projection decreases performance (74.22 $\rightarrow$ 68.40), confirming that our model better preserves instruction-relevant semantic structure compared to traditional dimensionality reduction techniques.

\paragraph{Generalization Across Embedding Models}
We test GSTransform on different generic embedding models (UAE, Mxbai, and BGE), and the results in Table \ref{tab:results} show consistent improvements (44.95 $\rightarrow$ 65.89, 45.60 $\rightarrow$ 66.01, and 43.48 $\rightarrow$ 65.32), verifying the method's robustness and model-agnostic adaptability.

\subsection{Parameter Studies}
\label{sec:expt:params}

To assess the robustness of GSTransform to key hyperparameters, we conduct a sensitivity analysis on three representative datasets--Toxic, AmzCF, and MultiHate.
Specifically, we vary: (1) the number of samples for transformation model training, and (2) the number of clusters ($k$) for label taxonomy. 

\begin{table}[t]
\small
\renewcommand{\arraystretch}{1.1}
\setlength\tabcolsep{4pt}
\centering
\resizebox{\columnwidth}{!}{
\begin{tabular}{lccccc}
  \toprule
  \multirow{2.5}{*}{\textbf{Dataset}} & \multicolumn{5}{c}{\textbf{\# Samples}} \\ \cmidrule{2-6} 
  & 1,000 & 2,000 & 3,000 & 4,000 & 5,000 \\
  \midrule
  \textbf{AmzCF (Clustering)} & 28.14 & 32.25 & 34.14 & 34.61 & 35.72 \\
  \textbf{MultiHate (STS)} & 49.04 & 50.65 & 52.45 & 54.13 & 54.64 \\
  \textbf{Toxic (Trip. Align.)} & 61.80 & 63.09 & 63.53 & 63.88 & 64.20 \\
  \textbf{Mean} & 46.33 & 48.66 & 50.04 & 50.87 & 51.52 \\
  \bottomrule
\end{tabular}}
\caption{The impact of the number of samples.}
\label{tab:sample}
\end{table}

\paragraph{Number of Samples}
We vary the number of training samples from 1,000 to 5,000 while keeping all other settings fixed.
Using Mxbai as the embedding backbone, we report the results in Table~\ref{tab:sample}.
The results show that while increasing the sample size generally leads to improved performance, the gains plateau around 3,000 samples, indicating diminishing returns beyond this point.
This finding highlights the efficiency of our method: it achieves strong performance with only a moderate amount of LLM-annotated data.

\begin{table}[t]
\small
\renewcommand{\arraystretch}{1.1}
\setlength\tabcolsep{4pt}
\centering 
\resizebox{\columnwidth}{!}{
\begin{tabular}{lccccc}
  \toprule
  \multirow{2.5}{*}{\textbf{Dataset}} & \multicolumn{5}{c}{\textbf{\# Clusters ($k$)}} \\ \cmidrule{2-6} 
  & 10 & 30 & 50 & 70 & 90 \\ 
  \midrule
  \textbf{AmzCF (Clustering)} & 35.37 & 33.95 & 34.14 & 34.87 & 33.46 \\
  \textbf{MultiHate (STS)} & 52.12 & 52.69 & 52.45 & 52.51 & 52.46 \\
  \textbf{Toxic (Trip. Align.)} & 63.59 & 64.17 & 63.53 & 64.05 & 63.31 \\
  \textbf{Mean} & 50.36 & 50.27 & 50.04 & 50.48 & 49.74 \\
  \bottomrule
\end{tabular}}
\caption{The impact of the number of clusters ($k$).}
\label{tab:k_value}
\end{table}

\paragraph{Number of Clusters}
We also evaluate the impact of varying the number of clusters $k$ in the $k$-means$++$ algorithm used for label construction. 
As depicted in Table \ref{tab:k_value}, performance remains largely stable across a wide range of $k$ values, suggesting that GSTransform is \emph{robust} to variations in label granularity. 
This insensitivity to hyperparameter tuning enhances its practicality in real-world settings, where optimal values of $k$ may be difficult to determine a priori.

\subsection{Case Study}
\label{sec:expt:case}

To showcase the effectiveness of GSTransform, we conduct a case study using the UAE embedding model on the NYTClust dataset, where users aim to cluster news articles by country.

\begin{figure}[ht]
 \centering
 \includegraphics[width=0.99\linewidth]{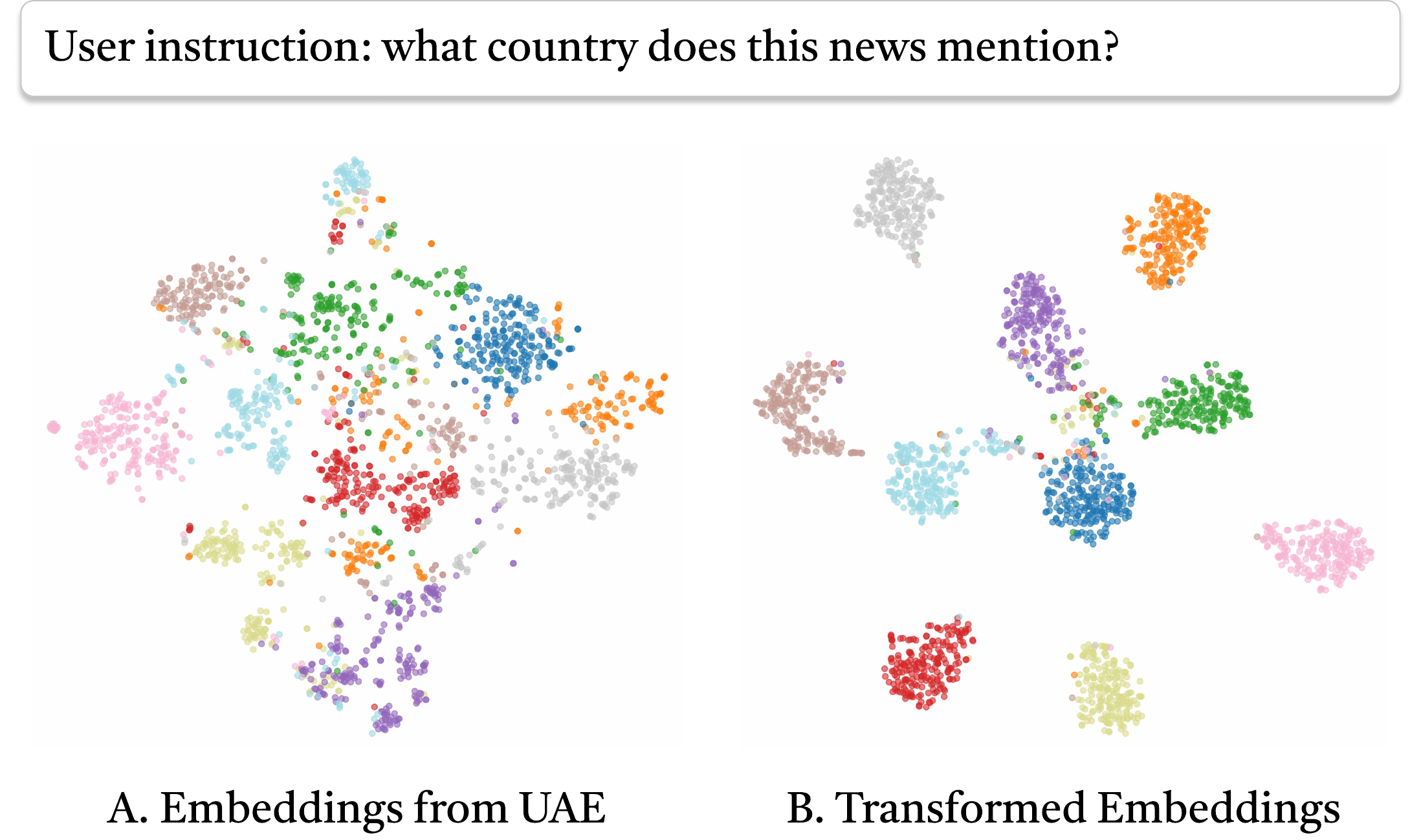}
 \vspace{-1.0em}
 \caption{The embedding visualization of UAE and GSTransform (UAE). We use t-SNE for projection and encode the country labels in different colors.}
 \label{fig:case}
\end{figure}

Consider a scenario where the user is interested in the location mentioned in the news. 
The user instructions can be ``What country does the news mention?''
Since the UAE model inherently does not support user instructions, the embedding results fail to clearly distinguish the news from different countries.
We visualized the UAE embeddings using t-SNE projection in Figure~\ref{fig:case}A.
Colors encode different ground-truth country labels.
The projection shows significant overlap between countries, with some news from the same country distributed in different areas.
Figure~\ref{fig:case}B visualizes the projection of the transformed embeddings based on user instruction.
The clusters are separated from each other more clearly and are more aggregated inside. It is easy to identify the number of clusters and the texts in each cluster, offering a clear overview of the countries in the news dataset.

\begin{figure}[ht]
 \centering
 \includegraphics[width=0.99\linewidth]{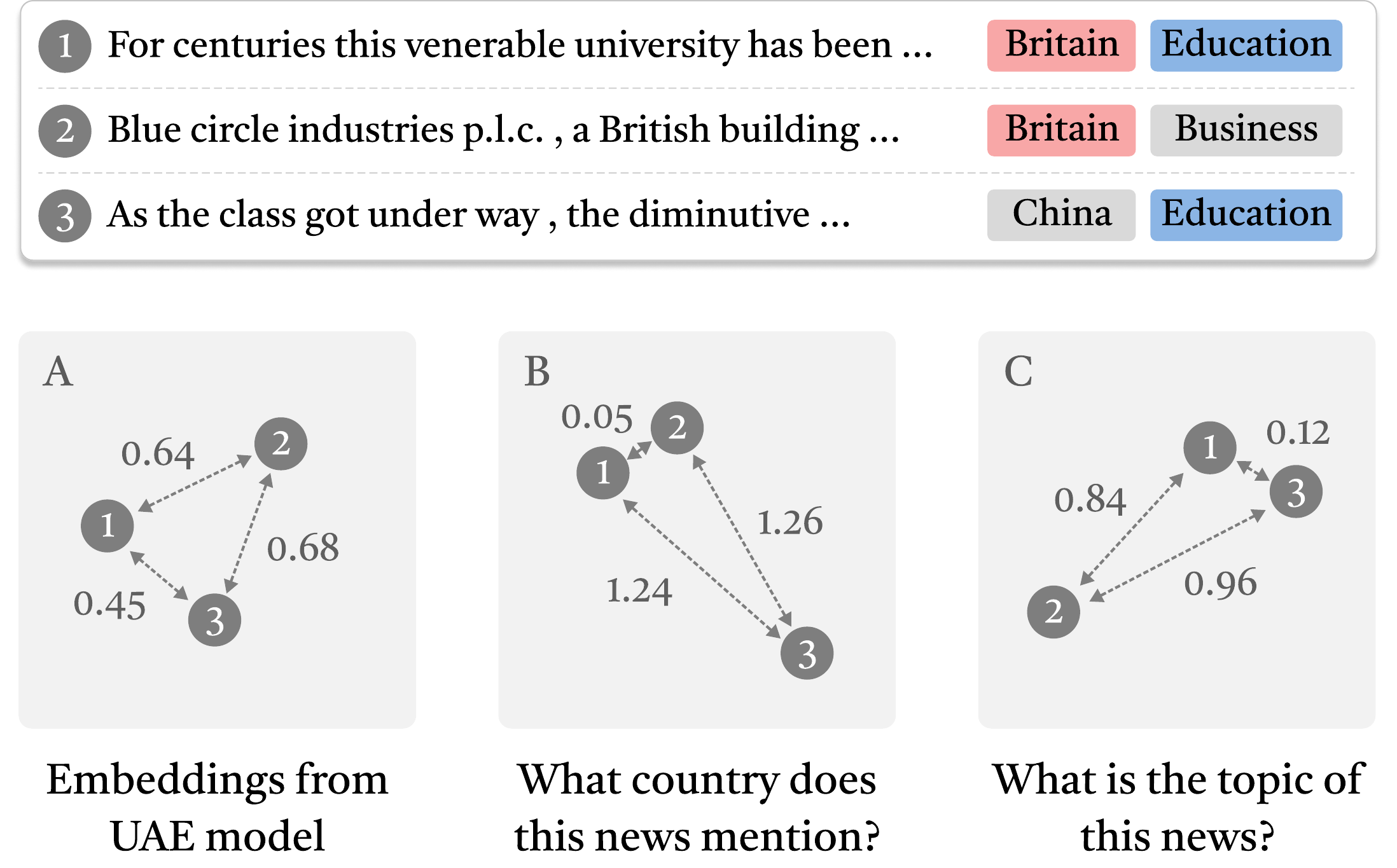}
 \vspace{-1.0em}
 \caption{The cosine distance relations between a triplet text based on different user instructions.}
 \label{fig:case-local}
\end{figure}

We also examine the cosine distance relation for a triplet text under different instructions in Figure~\ref{fig:case-local}.
In the triplet, the first text reports the same country (i.e., Britain) as the second text and has the same topic (i.e., education) as the third one.
Our method transforms the embeddings from the UAE model (Figure~\ref{fig:case-local}A) based on different user instructions.
When the user focuses on the countries of the news, the first and second texts are much closer (cosine distance = 0.05) as they share the same country (Figure~\ref{fig:case-local}B).
When the instruction focus shifts to the topic of the news, the first and third texts are closer (cosine distance = 0.12) due to their common topic of education (Figure~\ref{fig:case-local}C).
These results show that our method effectively adapts embeddings to reflect the user's focus, providing suitable distance relations that align with the instruction focus.

\section{Conclusions}
\label{sec:conclusions}

In this paper, we introduce GSTransform, a novel and efficient framework for generating instruction-following text embeddings via guided space transformation. 
Unlike prior methods that require full corpus re-encoding for every new instruction, GSTransform enables lightweight, real-time adaptation of pre-computed embeddings, significantly reducing computational overhead.
The framework consists of two key components: 
(1) Instruction-based Label Construction, which summarizes a sampled corpus based on user instructions and clusters the resulting embeddings to capture instruction-relevant semantics;
(2) Label-guided Embedding Transformation, which trains a simple encoder-decoder model to project generic embeddings into an instruction-aware semantic space using the constructed labels.
Extensive experiments across nine real-world datasets and three tasks demonstrate that GSTransform consistently improves embedding quality and outperforms state-of-the-art instruction-following baselines.
Notably, it achieves 6$\sim$300$\times$ speedups in real-time processing on large datasets, underscoring its scalability and efficiency.
Ablation and parameter studies further validate the effectiveness and robustness of each component.
GSTransform offers a promising path toward flexible, efficient, and instruction-adaptable embeddings, unlocking new opportunities for user-customized semantic representations in real-world applications.

\section*{Limitations}
\label{sec:limitations}

\paragraph{Robustness of Transformation Quality} 
The robustness of transformation quality is influenced by several factors.
First, while random text sampling for label construction is effective, it can be sensitive to data imbalance, underscoring the potential of advanced methods like coreset selection to enhance data representativeness \cite{coresetkdd23}.
Second, the capability of generic embeddings to capture instruction-specific nuances is critical, highlighting the need for high-quality generic embeddings as a reliable foundation for transformation.

\paragraph{Encoder-Decoder Model Design}
The current encoder-decoder architecture utilizes simple linear layers, striking a balance between performance and computational efficiency. Although this design choice is sufficient for our current objectives, future research could investigate more sophisticated architectural variants, such as incorporating non-linear transformations or attention mechanisms.
These extensions should carefully weigh potential performance gains against increased computational complexity.

\section*{Acknowledgments}
This research is supported by 
the National Natural Science Foundation of China under grant Nos. 62132017, 62302435,
the ``Pioneer'' and ``Leading Goose'' R\&D Program of Zhejiang (2024C01167),
Zhejiang Provincial Natural Science Foundation of China under Grant No. LD24F020011,
the Ministry of Education, Singapore, under its MOE AcRF TIER 3 Grant (MOE-MOET32022-0001),
and the National Research Foundation, Singapore, under its AI Singapore Programme (AISG Award No.~AISG3-RP-2022-029).
Any opinions, findings, and conclusions or recommendations expressed in this material are those of the author(s) and do not reflect the views of the Ministry of Education, Singapore, or the National Research Foundation, Singapore.

\bibliography{acl_main}
\appendix

\section{Prompt Templates}
\label{appendix:prompts}

\subsection{Text Summarization}
\label{appendix:prompts:text_summ}

This prompt instructs the LLM to generate a concise summary based on a given instruction. It emphasizes extracting key points without unnecessary details and also specifies the output format at the end of the prompt to facilitate result parsing.

\begin{tcolorbox}[colback=white]
Summarize the text based on the following instruction. The summary must focus on the instruction's key points and not exceed 10 words.\\
\textbf{Instruction}: \textit{\{instruction\}}\ \\
\textbf{Text}: \textit{\{text\}} \\
\textbf{Required Format}: Summary: <summary>\\
\textbf{Note}: Only output the summary in English starting with "Summary:", do not include any other text.
\end{tcolorbox}

\begin{table*}[ht]
\small
\renewcommand{\arraystretch}{1.3}
\setlength\tabcolsep{4pt}
\centering
\resizebox{\textwidth}{!}{
\begin{tabular}{llll}
  \toprule
  \textbf{Downstream Task} & \textbf{Dataset} & \textbf{Instruction Focus} & \textbf{Category} \\
  \midrule
  \multirow{5}{*}{Clustering} & \multirow{2.3}{*}{NYTClust} & Topic & \{Estate, Technology, Science, $\cdots$, Sports\} \\ \cmidrule{3-4}  
  &  & Location & \{Germany, France, America, China, Canada, Britain, $\cdots$, Japan\} \\ \cmidrule{2-4} 
  & AmzCF & Counterfactual & \{Counterfactual, Not-Counterfactual\} \\ \cmidrule{2-4} 
  & MNews & Topic & \{Business, Politics, Sports, Entertainment, Health, $\cdots$, Technology\} \\
  \midrule
  \multirow{6.5}{*}{STS} & \multirow{2.3}{*}{PaperCode} & Method & \{Gaussian Process, Q-Learning, SVM, CLIP, PCA, $\cdots$, NeRF\} \\ \cmidrule{3-4} 
  &  & Task & \{Federated Learning, Active Learning, $\cdots$, Question Answering\} \\ \cmidrule{2-4} 
  & \multirow{2}{*}{MultiHate} & Hateful & \{Hateful, Non-Hateful\} \\ \cmidrule{3-4} 
  &  & Language & \{Arabic, German, French, Italian, $\cdots$, Mandarin\} \\ \cmidrule{2-4} 
  & Big Patent & Patent Category & \{Human Necessities, Physics, Electricity, Chemistry, $\cdots$, Textiles\} \\
  \midrule
  \multirow{5}{*}{Triplet Alignment} & \multirow{2.3}{*}{IntEmo} & Intent & \{Track Card, Inquiry Exchange Rates, $\cdots$, Cash Withdrawal\} \\ \cmidrule{3-4} 
  &  & Emotion & \{Positive, Negative\} \\ \cmidrule{2-4} 
  & Toxic & Toxic & \{Toxic, Not Toxic\} \\ \cmidrule{2-4} 
  & AG-News & Topic & \{World, Sports, Business, Sci/Tech\} \\
  \bottomrule
\end{tabular}}
\caption{The downstream tasks and their associated datasets. This benchmark involves diverse semantic aspects that can be specified in the user instructions, including topic, location, language, and intent.}
\label{tab:tasks}
\end{table*}

\subsection{Label Generation}
\label{appendix:prompts:label_gene}

This prompt instructs the LLM to define a clear, precise category label that captures the core feature of the positive texts while distinguishing them from negative texts. It also specifies the output format at the end of the prompt to facilitate result parsing.

\begin{tcolorbox}[colback=white]
Analyze these two groups of texts and define a clear category label that best describes the characteristics of the current group based on the following instructions.\\
\textbf{Instruction}: \textit{\{instruction\}}\\
\textbf{Current Group Texts}: \textit{\{positive\_texts\}}\\
\textbf{Other Group Texts}: \textit{\{negative\_texts\}}\\
\textbf{Requirements}:\\
- The label MUST strictly follow and reflect the given instruction.\\
- Focus on the main characteristics of the current group based on the instruction.\\
- Label should be generalizable but distinguishable from other texts.\\
- Use clear and precise language.\\
- The category name should be no more than 5 words.\\
\textbf{Required Format}: Category: <category>\\
\textbf{Note}: Only output the category name starting with "Category:", do not include any other text.
\end{tcolorbox}

\subsection{Text Classification}
\label{appendix:prompts:text_classf}

This prompt instructs the LLM to classify a given text based on the user instruction and predefined categories. It also specifies the output format at the end of the prompt to facilitate result parsing.

\begin{tcolorbox}[colback=white]
Please classify the following text based on the instruction and available categories.

\textbf{Instruction}: \textit{\{instruction\}} \\
\textbf{Available Categories}: \textit{\{categories\}} \\
\textbf{Text to Classify}: \textit{\{text\}} \\
\textbf{Required Format}: Classification: <category\_name> \\  
\textbf{Note}: Only output the category name starting with "Classification:", do not include any other text. The category must be exactly as listed above.
\end{tcolorbox}

\section{Downstream Tasks and Datasets}
\label{appendix:tasks_datasets}

Table~\ref{tab:tasks} summarizes the downstream tasks and datasets used to evaluate instruction-aware embedding performance.
Our benchmark spans three representative tasks, i.e., clustering, semantic textual similarity (STS), and triplet alignment, which are designed to assess how well embeddings reflect user-specified semantic intents.

The datasets cover a wide range of instruction-relevant aspects, including topic categories (e.g., AG-News, NYTClust), geographic locations (e.g., MNews), language types (e.g., MultiHate), and subjective attributes such as intent and emotion (e.g., IntEmo).
These diverse tasks and semantic dimensions allow for a rigorous and comprehensive evaluation of how effectively GSTransform adapts pre-trained embeddings to align with different instruction-driven perspectives.

\section{Dataset-oriented Instructions}
\label{appendix:instructions}

\begin{table*}[]
\resizebox{\textwidth}{!}{
\renewcommand{\arraystretch}{1.2}
\begin{tabular}{llll}
\toprule
\textbf{Dataset} & \textbf{Aspect} & \textbf{Model} & \textbf{Instruction} \\ \midrule
\multirow{6.25}{*}{NYTClust} & \multirow{3}{*}{Topic} & InstructOR & Represent the text based on the main news category. \\
 &  & InBedder & What is the main category of this news? \\
 &  & Ours & What is the main category of this news? \\ \cmidrule(lr){2-4}
 & \multirow{3}{*}{Location} & InstructOR & Represent the text based on where the news happen. \\
 &  & InBedder & Where did the news happen? \\
 &  & Ours & What country does this news mention? Just tell me the country name. \\ \midrule
\multirow{3}{*}{AmzCF} & \multirow{3}{*}{Counterfactual} & InstructOR & Represent the text based on whether it is counterfactual (yes/no). \\
 &  & InBedder & Is the sentence counterfactual? \\
 &  & Ours & Is the sentence counterfactual? just tell me yes/no. \\ \midrule
\multirow{3}{*}{MNews} & \multirow{3}{*}{Topic} & InstructOR & Represent the text based on the main category of the news. \\
 &  & InBedder & What is the main category of this news? \\
 &  & Ours & What is the main category of this news? \\ \midrule
\multirow{6.25}{*}{PaperCode} & \multirow{3}{*}{Method} & InstructOR & Represent the paper based on the method used. \\
 &  & InBedder & What is the method used in this paper? \\
 &  & Ours & What is the method used in this paper? \\ \cmidrule(lr){2-4}
 & \multirow{3}{*}{Task} & InstructOR & Represent the paper abstract based on the research task. \\
 &  & InBedder & What is the research task of this paper abstract? \\
 &  & Ours & What is the research task of this paper abstract? \\ \midrule
\multirow{6.25}{*}{MultiHate} & \multirow{3}{*}{Hateful} & InstructOR & Represent the text based on whether it is hateful or not. \\
 &  & InBedder & Is the text hateful? Just tell me yes/no. \\
 &  & Ours & Is the text hateful? Just tell me yes/no. \\ \cmidrule(lr){2-4}
 & \multirow{3}{*}{Language} & InstructOR & Represent the text based on the language. \\
 &  & InBedder & What is the language of the text? Just tell me the language. \\
 &  & Ours & What is the language of the text? Just tell me the language. \\ \midrule
\multirow{5}{*}{Big Patent} & \multirow{5}{*}{Patent Category} & InstructOR & \begin{tabular}[c]{@{}l@{}}Represent the text based on the category the patent belongs to according\\  to the Cooperative Patent Classification (CPC) code.\end{tabular} \\
 &  & InBedder & \begin{tabular}[c]{@{}l@{}}What category does the patent belong to according to the Cooperative \\ Patent Classification (CPC) code?\end{tabular} \\
 &  & Ours & \begin{tabular}[c]{@{}l@{}}What category does the patent belong to according to the Cooperative\\ Patent Classification (CPC) code?\end{tabular} \\ \midrule
\multirow{6.25}{*}{IntEmo} & \multirow{3}{*}{Intent} & InstructOR & Represent the text based on what the customer needs. \\
 &  & InBedder & What does the customer need? \\
 &  & Ours & What does the user care about? Just tell me the name of the thing. \\ \cmidrule(lr){2-4}
 & \multirow{3}{*}{Emotion} & InstructOR & Represent the text based on the emotion of the user. \\
 &  & InBedder & What is the emotion of the user? \\
 &  & Ours & What is the emotion of the user? Just tell me the emotion. \\ \midrule
\multirow{3}{*}{Toxic} & \multirow{3}{*}{Toxic} & InstructOR & Represent the text based on whether it is toxic (yes/no). \\
 &  & InBedder & Is the sentence toxic? just tell me yes/no. \\
 &  & Ours & Is the sentence toxic? just tell me yes/no. \\ \midrule
\multirow{3}{*}{AG-News} & \multirow{3}{*}{Topic} & InstructOR & Represent the news according to their topic category. \\
 &  & InBedder & What is the topic category of this news? \\
 &  & Ours & What is the topic category of this news? Just tell me the category name. \\ \bottomrule
\end{tabular}}
\caption{Instructions of InstructOR, InBedder, and GSTransform for each dataset.}
\label{tab:instruction}
\end{table*}

We provide the task instructions used by InstructOR, InBedder, and GSTransform for each dataset in Table~\ref{tab:instruction}.
The baseline instructions adhere to their default stylistic conventions: InstructOR uses command-style prompts, while InBedder adopts a question-style format. Despite these stylistic differences, both formats convey the same semantic intent.

The instructions used in GSTransform largely mirror those of InBedder, with only minor deviations, primarily related to preferred response length, which are implicitly controlled through system-level prompts in InBedder.
To ensure fair comparison and model consistency, we standardize these differences while preserving semantic equivalence across all methods.

\section{Full Results of Ablation Studies}
\label{app:full_ablation}

\begin{table*}[ht]
\small
\renewcommand{\arraystretch}{1.5}
\setlength\tabcolsep{4pt}
\centering 
\resizebox{\linewidth}{!}{
\begin{tabular}{lcccccccccc}
  \toprule
  \multirow{2.5}{*}{\textbf{Solutions}} 
  & \multicolumn{3}{c}{\textbf{Clustering}} & \multicolumn{3}{c}{\textbf{STS}} & \multicolumn{3}{c}{\textbf{Triplet Alignment}} & \multirow{2.5}{*}{\textbf{Mean $\uparrow$}} \\ \cmidrule(lr){2-4} \cmidrule(lr){5-7} \cmidrule(lr){8-10}
  & \textbf{NYTClust} & \textbf{AmzCF} & \textbf{MNews} & \textbf{PaperCode} & \textbf{MultiHate} & \textbf{Big Patent} & \textbf{IntEmo} & \textbf{Toxic} & \textbf{AG-News} &  \\ \midrule
  \textbf{GSTransform (Mxbai)} & 73.92 & 34.14 & 64.71 & 83.43 & 52.45 & 38.34 & 96.30 & 63.53 & 87.27 & 66.01 \\
  \textbf{Remove Summ.} & 68.32 & 38.85 & 64.14 & 81.72 & 16.21 & 34.81 & 85.14 & 57.55 & 86.47 & 59.25 \\
  \textbf{Directed Label Gen.} & 42.50 & 14.60 & 18.24 & 30.04 & 20.55 & 26.97 & 86.76 & 55.06 & 74.30 & 41.00 \\
  \textbf{FDA-based Transf.} & 67.53 & 35.77 & 63.15 & 75.15 & 48.49 & 21.56 & 89.19 & 57.81 & 74.10 & 59.19 \\
  \bottomrule
\end{tabular}}
\caption{Full results of performance evaluations for different solutions in ablation studies.}
\label{tab:ablation_all}
\end{table*}

Table~\ref{tab:ablation_all} presents the full results of our ablation studies, evaluating the contribution of each core component in the GSTransform framework across all datasets.
We assess the impact of (1) removing the instruction-based summarization step, (2) replacing the full label construction process with direct LLM-generated labels, and (3) substituting our transformation model with an FDA baseline.

These results offer a more comprehensive view of how each design choice affects model performance, reinforcing the necessity of both the instruction-based label construction and the label-guided embedding transformation for achieving robust instruction-following behavior.

\begin{table}[h!]
\centering
\renewcommand{\arraystretch}{1.3}
\setlength\tabcolsep{4pt}
\resizebox{\linewidth}{!}{%
\begin{tabular}{rrrr}
\toprule
\textbf{Data Size} & \textbf{InstructOR} & \textbf{InBedder-Roberta} & \textbf{GSTransform} \\ 
\midrule
1,000              & 12 s          & 7 s             & 36 s     \\ 
3,000              & 28 s          & 19 s            & 68 s     \\ 
10,000             & 74 s          & 60 s            & 68 s     \\ 
30,000             & 173 s         & 175 s           & 69 s     \\ 
100,000            & 450 s         & 583 s           & 70 s     \\ 
\bottomrule
\end{tabular}}
\caption{Real-time latency comparison across varied dataset sizes on AG-News.}
\label{tab:efficiency_ag_news}
\end{table}

\begin{table}[h!]
\centering
\renewcommand{\arraystretch}{1.2}
\setlength\tabcolsep{4pt}
\resizebox{\linewidth}{!}{%
\begin{tabular}{rrrr}
\toprule
\textbf{Data Size} & \textbf{InstructOR} & \textbf{InBedder-Roberta} & \textbf{GSTransform} \\ 
\midrule
1,000        & 36 s          & 27 s          & 39 s      \\ 
3,000        & 95 s          & 81 s          & 83 s      \\ 
10,000       & 291 s         & 266 s         & 86 s      \\ 
30,000       & 808 s         & 794 s         & 85 s      \\ 
50,000       & 1,304 s       & 1,327 s       & 87 s      \\ 
\bottomrule
\end{tabular}}
\caption{Real-time latency comparison across varied dataset sizes on Big Patent.}
\label{tab:efficiency_big_patent}
\end{table}

\section{Efficiency Comparisons across Different Dataset Scales}
\label{app:efficiency_comparison}

We compare the real-time latency of GSTransform and baseline models across varying dataset sizes. 
As shown in Tables~\ref{tab:efficiency_ag_news} and \ref{tab:efficiency_big_patent}, baseline latency scales linearly with data volume, making them more suitable for small-scale scenarios. 
For instance, they remain competitive on the Big Patent dataset when the size is under 3,000 samples. 
In contrast, GSTransform achieves near-constant latency regardless of dataset size, as its transformation relies on a fixed-size sample. This scalability results in substantial efficiency advantages for large-scale applications.

\end{document}